\documentclass[runningheads]{llncs}
\usepackage[T1]{fontenc}
\usepackage{graphicx}
\usepackage{booktabs}
\usepackage[misc]{ifsym}

\usepackage{amsmath,amssymb,amsfonts}
\usepackage{algorithmic}
\usepackage{textcomp}
\usepackage{hyperref}
\usepackage{float}
\usepackage{enumitem}
\usepackage{subcaption}
\usepackage{xcolor}
\usepackage{xfrac}
\usepackage{longtable}
\usepackage{makecell}
\usepackage{rotating}
\newcommand{\samethanks}[1][\value{footnote}]{\footnotemark[#1]}
\newcommand{\weaseltwo}{Sch\"afer and Leser}
\renewcommand{\ackname}{Acknowledgements.}

\begin{document}

\title{Revisiting WEASEL~2.0: Reproduction, Sensitivity, and an Adaptive Ensemble-Size Rule}

\titlerunning{Revisiting WEASEL~2.0}

\author{Cian Higgins\thanks{Equal contribution.} \and Gerard Carrigan\samethanks \and Pinar Sungu Isiacik \and Georgiana Ifrim}

\authorrunning{C. Higgins et al.}

\institute{University College Dublin, Dublin, Ireland \\
\email{\{cian.higgins1,gerard.carrigan1,pinar.sunguisiacik\}@ucdconnect.ie}
\email{georgiana.ifrim@ucd.ie}}

\maketitle

\begin{abstract}
WEASEL~2.0 is a dictionary-based time series classifier that combines dilated sliding windows with a randomised hyperparameter ensemble and a fixed-size dense feature representation. Two of its hyperparameter choices, the maximum ensemble size and the maximum window size, are specified by simple thresholding rules whose chosen thresholds are not empirically justified in the original paper. In this work we reproduce WEASEL~2.0 on 114 UCR datasets, achieving a mean accuracy of 0.865 and median of 0.928, closely matching the published values (Wilcoxon signed-rank, $p = 0.655$). We then test the sensitivity of four design choices: the downstream classifier, the absence of feature weighting, the maximum window-size rule, and the maximum ensemble-size rule. The first three are robust to perturbation. The fourth is over-provisioned for long-series datasets, motivating an adaptive rule that sets the maximum ensemble size from series length and number of classes. Evaluated on fixed-length datasets, the adaptive rule reduces peak fit memory by a median of 37\,MB (mean 395\,MB) and fit time by a median of 0.4\,s (mean 4\,s), with a median accuracy change of 0\% (mean $-0.11$\%). Memory and time savings concentrate on long-series datasets where the original rule allocates the largest ensemble size.

\keywords{WEASEL~2.0 \and time series classification \and dictionary methods \and reproducibility \and UCR benchmark}
\end{abstract}

\section{Introduction}
\label{sec:intro}

Time series classification (TSC) assigns a discrete label to an ordered sequence of real values, with applications spanning medical diagnostics, industrial monitoring, and activity recognition. The field has matured rapidly, with methods broadly falling into distance-based, dictionary-based, kernel-based, shapelet-based, deep learning, and hybrid categories~\cite{bagnall2017bakeoff,middlehurst2024bakeoff}.

Dictionary-based methods such as BOSS~\cite{schaefer2015boss}, TDE~\cite{middlehurst2021tde}, and WEASEL~\cite{schaefer2017weasel} extract symbolic patterns from time series using sliding windows and the Symbolic Fourier Approximation (SFA)~\cite{schaefer2012sfa}. While historically competitive, these methods suffered from unpredictable memory consumption and have been surpassed in accuracy by kernel methods like ROCKET~\cite{dempster2020rocket} and its successors MiniRocket~\cite{dempster2021minirocket} and MultiRocket~\cite{tan2022multirocket}.

WEASEL~2.0~\cite{weasel2.0} addresses two design issues in WEASEL~1.0~\cite{schaefer2017weasel} and other dictionary methods:
excessive memory consumption and sensitivity to minor subsequence changes. It introduces a dilation mapping to increase the receptive field of sliding windows, and a fixed-size dense feature representation combining a compact 256-word dictionary, variance-based Fourier coefficient selection, and an ensemble of randomised hyperparameter configurations, producing a controlled feature vector of roughly 20k--70k features. On the UCR benchmark, the authors report that WEASEL~2.0 is the most accurate dictionary classifier and not significantly different from the best non-ensemble methods such as MultiRocket~\cite{tan2022multirocket} and R-DST~\cite{guillaume2022rdst}. Two heuristics in the original paper, governing the maximum ensemble size and the maximum
window size, are presented as rules of thumb without empirical justification for the chosen thresholds. No independent reproduction of WEASEL~2.0 has been published, and the sensitivity of its design choices to perturbation has not been examined.

In this work, we first ask whether the published WEASEL~2.0 results reproduce under independent analysis. We then examine whether the under-justified design choices are empirically robust to perturbation, and finally whether the ensemble-size heuristic can be refined to reduce resource use without accuracy loss. Our contributions are:

\begin{itemize}
    \item An independent reproduction of WEASEL~2.0, confirming published accuracy on 114 UCR datasets.
    \item A sensitivity analysis of four under-justified design choices, three of which we find empirically robust.
    \item A diagnostic finding that the ensemble-size heuristic is over-provisioned for long-series datasets.
    \item An adaptive ensemble-size rule based on series length and class count that reduces peak fit memory at minimal accuracy cost.
\end{itemize}

\section{Related Work}
\label{sec:related}

Dictionary-based classifiers build symbolic representations from sliding windows and classify based on pattern frequencies. Early methods such as BOSS~\cite{schaefer2015boss} established the strength of this approach but also exposed a key limitation: large and sometimes unpredictable memory usage. Subsequent work focused on making these models more scalable or more accurate by modifying the transform or the ensemble construction rather than redesigning the entire classifier family. cBOSS and related scalable dictionary variants reduce training cost by randomising ensemble selection and restricting the number of candidate models, while still maintaining accuracy~\cite{Middlehurst_2019_cboss}. TDE~\cite{middlehurst2021tde} introduces temporal sensitivity to the BOSS framework. WEASEL~\cite{schaefer2017weasel} improved over earlier dictionary methods through supervised symbolic feature generation, but its sparse feature space could still become very large. WEASEL~2.0~\cite{weasel2.0} addresses this through dilation, a compact fixed-size dictionary, and randomised hyperparameter ensembling, producing a dense representation with controlled memory consumption.

A similar pattern of refining an established design rather than introducing a new one is visible across the TSC literature. ROCKET~\cite{dempster2020rocket} introduced random convolutional kernels with a simple linear classifier. MiniRocket~\cite{dempster2021minirocket} and MultiRocket~\cite{tan2022multirocket} refined this design to improve runtime and accuracy. R-DST~\cite{guillaume2022rdst} extends shapelet-based classification using random dilation to strengthen accuracy while maintaining scalability. Hydra~\cite{dempster2023hydra} hybridises kernel and dictionary approaches by introducing competing convolutional kernels. These methods show that substantial gains can be obtained by revisiting the design of an existing classifier rather than introducing a completely new model class.

Empirical evaluation of TSC methods on the UCR archive has been standardised through the bake-off studies of Bagnall et al.~\cite{bagnall2017bakeoff} and Middlehurst et al.~\cite{middlehurst2024bakeoff}. These studies provide cross-classifier comparisons under a unified protocol, including WEASEL~2.0 and the kernel and shapelet methods we consider. They do not, however, examine the sensitivity of individual classifiers to their internal design choices, nor do they verify reproducibility of the reported numbers under an independent analysis. This paper contributes to both: we reproduce WEASEL~2.0 on the same 114-dataset subset of the UCR archive used in the original paper, and we explicitly test the robustness of its design choices to perturbation.

Unlike the works above that propose new method designs, our paper takes the complementary approach of empirically examining the design choices of an existing classifier and refining one of its heuristics. We treat WEASEL~2.0 as a fixed reference and ask which of its components contribute to its accuracy, how robust its heuristic thresholds are, and where a simple data-aware refinement can reduce resource use without significant accuracy loss.

\section{Reproducing WEASEL~2.0}
\label{sec:reproduction}

The authors of WEASEL~2.0 provide the source code on GitHub\footnote{\href{https://github.com/patrickzib/dictionary}{https://github.com/patrickzib/dictionary}} and it is also available in the aeon toolkit~\cite{middlehurst2024aeon}. WEASEL~2.0 uses RidgeClassifierCV from scikit-learn~\cite{scikit-learn} as the downstream linear model. We cloned the GitHub repository and used aeon to access the UCR benchmark~\cite{ucr2018} dataset. Our reproduction\footnote{\href{https://github.com/gerryc-0/Why-so-Time-Serious}{https://github.com/gerryc-0/Why-so-Time-Serious}} used Python 3.12, scikit-learn 1.6.1, and aeon 1.3.0, executed on a machine with an AMD Ryzen AI 7 350 with Radeon 860M (2.00\,GHz) processor and 16\,GB RAM running Windows 11. We used 4 threads for parallelism and the random seed $1379$ everywhere to match the authors' configuration.

We ran WEASEL~2.0 using the source code provided on the same 114 UCR datasets reported in the original paper. Our script instantiates the WEASEL~2.0 classifier with default parameters, fits on the training split, and records test accuracy, fit time, predict time, and total feature count. To verify the paper's claims when comparing to other classifiers, we also reproduced results for seven other classifiers: MultiRocket~\cite{tan2022multirocket}, MiniRocket~\cite{dempster2021minirocket}, ROCKET~\cite{dempster2020rocket}, R-DST~\cite{guillaume2022rdst}, Hydra~\cite{dempster2023hydra}, WEASEL~1.0~\cite{schaefer2017weasel}, and cBOSS~\cite{Middlehurst_2019_cboss}. All classifiers used the same random seed and number of threads as seen above in the WEASEL~2.0 configuration. A small number of datasets for WEASEL~1.0 initially crashed due to memory issues but ran successfully after reducing the number of threads to one (for Crop, ElectricDevices, and FordB). All other classifiers ran successfully on all 114 datasets. \weaseltwo~\cite{weasel2.0} stored the UCR archive locally in their original experiments; in our reproduction, we loaded the datasets directly using the aeon library (\texttt{load\_classification}).

Table~\ref{tab:reproduction} compares our WEASEL~2.0 reproduction with the paper's published values. Our mean accuracy of 0.865 and median of 0.928 closely match the reported 0.863 and 0.927, respectively.

\begin{table}[ht!]
\caption{WEASEL~2.0 reproduction vs.\ the paper.}
\label{tab:reproduction}
\centering
\begin{tabular}{lcc}
\toprule
 & Paper & Ours \\
\midrule
Mean accuracy      & 0.863  & 0.865 \\
Median accuracy    & 0.927  & 0.928 \\
Datasets tested    & 114    & 114 \\
Total fit time     & 1.71\,h & 0.57\,h \\
Total predict time & 1.79\,h & 0.69\,h \\
\bottomrule
\end{tabular}
\end{table}

When comparing per-dataset accuracies against the paper's published values on all 114 datasets, 76 (66.7\%) are within a tolerance of $\pm 0.1\%$, which we consider effectively reproduced. Of the remaining 38 datasets, our reproduction scores were higher on 16 and lower on 22 datasets. A Wilcoxon signed-rank test~\cite{wilcoxon} confirms no statistically significant difference between our results and the published values ($p = 0.655$). Nine datasets differ by more than 1\%. The largest are PickupGestureWiimoteZ ($+12.0\%$) and ShakeGestureWiimoteZ ($+8.0\%$), both variable-length gesture datasets. The original WEASEL~2.0 paper does not specify how these datasets were transformed to equal-length series, and the aeon implementation we use truncates them to a fixed length. We were unable to verify whether \weaseltwo~\cite{weasel2.0} used the same strategy. Among the standard fixed-length datasets, the largest outliers are ACSF1 ($-2.0\%$) and ScreenType ($+1.9\%$).

Table~\ref{tab:competitors} shows our reproduced accuracy for all eight classifiers. WEASEL~2.0 achieves the highest median accuracy (0.928), consistent with the paper's boxplot~\cite{weasel2.0}. MultiRocket has the highest mean accuracy (0.867), with WEASEL~2.0 and R-DST close behind (both 0.865). WEASEL~2.0 outperforms WEASEL~1.0 (80 wins vs.\ 18 losses) and cBOSS (81 wins vs.\ 19 losses), confirming the claim that WEASEL~2.0 is the strongest dictionary classifier.

\begin{table}[ht!]
\caption{Reproduced accuracy summary for all classifiers (114 UCR datasets). D = Dictionary, K = Kernel, S = Shapelet.}
\label{tab:competitors}
\centering
\begin{tabular}{lcccc}
\toprule
Classifier & Type & Mean & Median & N \\
\midrule
WEASEL~2.0   & D   & 0.865 & \textbf{0.928} & 114 \\
MultiRocket  & K   & \textbf{0.867} & 0.918 & 114 \\
MiniRocket   & K   & 0.861 & 0.909 & 114 \\
ROCKET       & K   & 0.857 & 0.908 & 114 \\
R-DST        & S   & 0.866 & 0.904 & 114 \\
Hydra        & K/D & 0.858 & 0.900 & 113 \\
WEASEL~1.0   & D   & 0.826 & 0.872 & 114 \\
cBOSS        & D   & 0.816 & 0.858 & 114 \\
\bottomrule
\end{tabular}
\end{table}

\begin{figure}[ht!]
\centering
\includegraphics[width=\columnwidth]{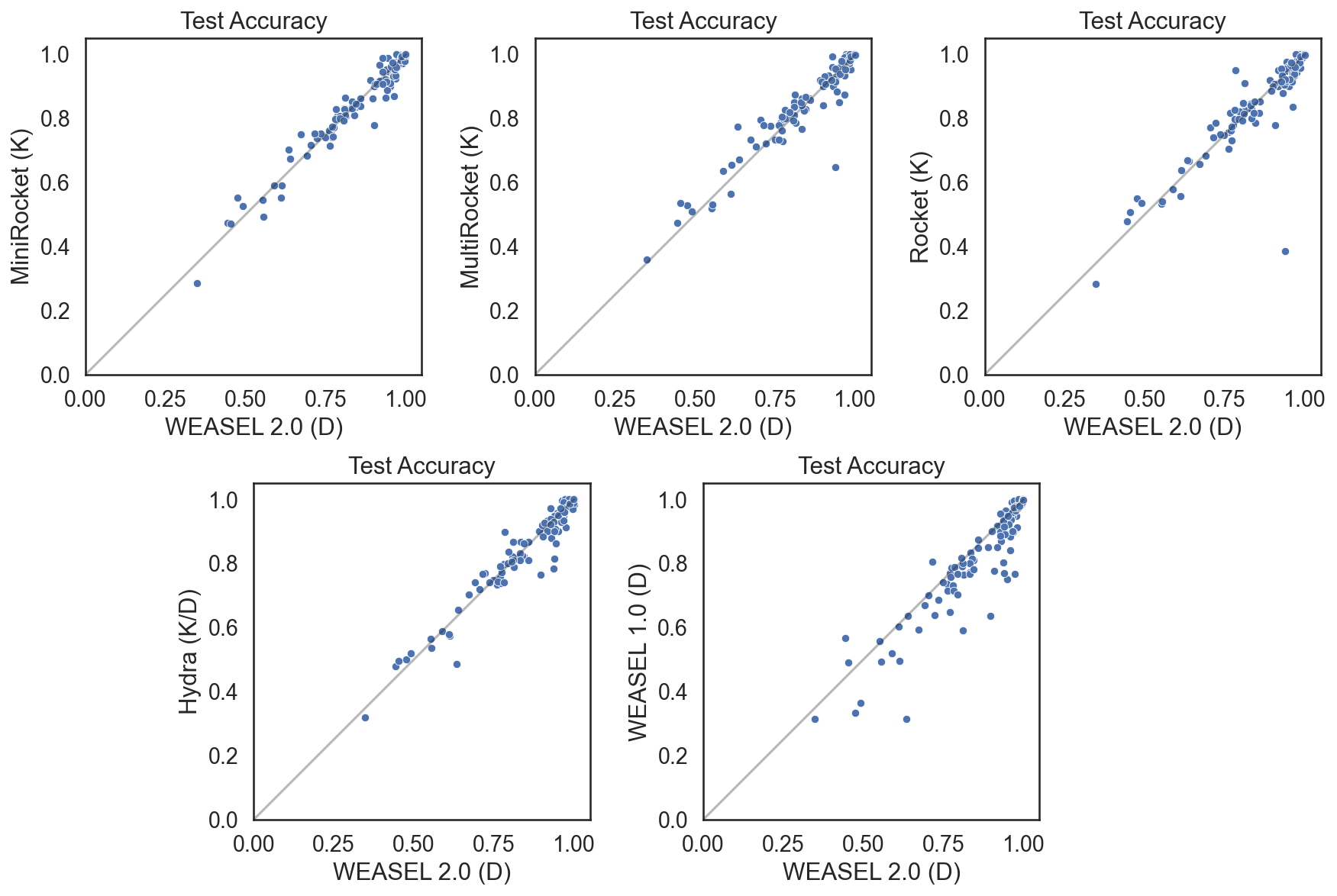}
\caption{Reproduced pairwise accuracy comparisons: WEASEL~2.0 (x-axis) vs.\ each competitor (y-axis). Points below the diagonal favour WEASEL~2.0.}
\label{fig:scatter_grid}
\end{figure}

Pairwise scatter plots in Figure~\ref{fig:scatter_grid} compare WEASEL~2.0's test accuracy against each competitor, with each point representing one dataset and the diagonal indicating equal performance. Against MiniRocket and MultiRocket, points cluster tightly along the diagonal, indicating comparable performance across the UCR archive. ROCKET follows a similar pattern but with slightly more variation at mid-range accuracies. For Hydra, points are concentrated in the upper-right region, reflecting high accuracy for both methods on most datasets, with a marginal advantage to WEASEL~2.0. The most asymmetric plot is against WEASEL~1.0, where the majority of points fall below the diagonal, indicating consistent improvement in WEASEL~2.0 across the full accuracy range. This suggests that the architectural changes of WEASEL~2.0, dilation and randomised ensembles, provide significant accuracy gains over the earlier version.

\begin{figure}[ht!]
\centering
\includegraphics[width=\columnwidth]{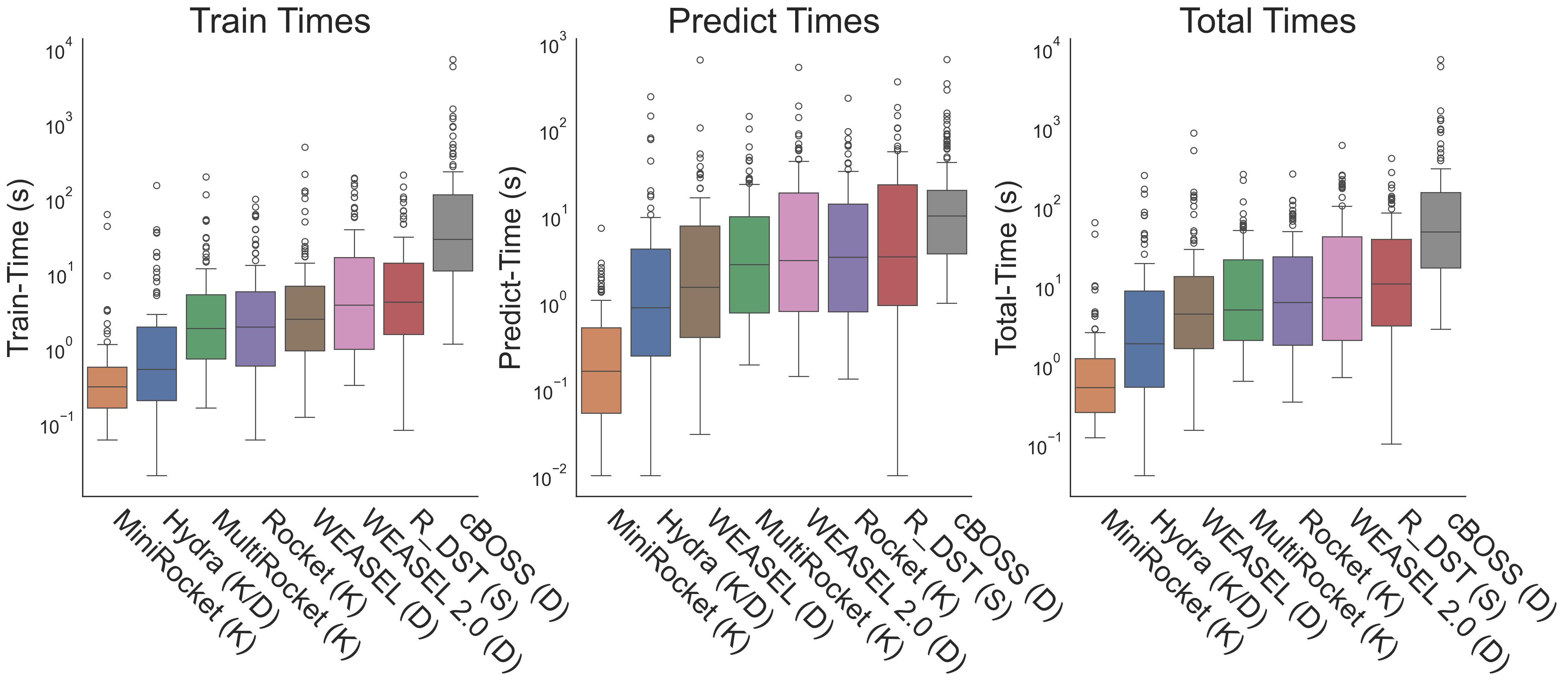}
\caption{Reproduced train, predict, and total runtime distributions (log scale) across fixed-length UCR datasets.}
\label{fig:runtime}
\end{figure}

\begin{figure}[ht!]
\centering
\includegraphics[width=0.8\textwidth]{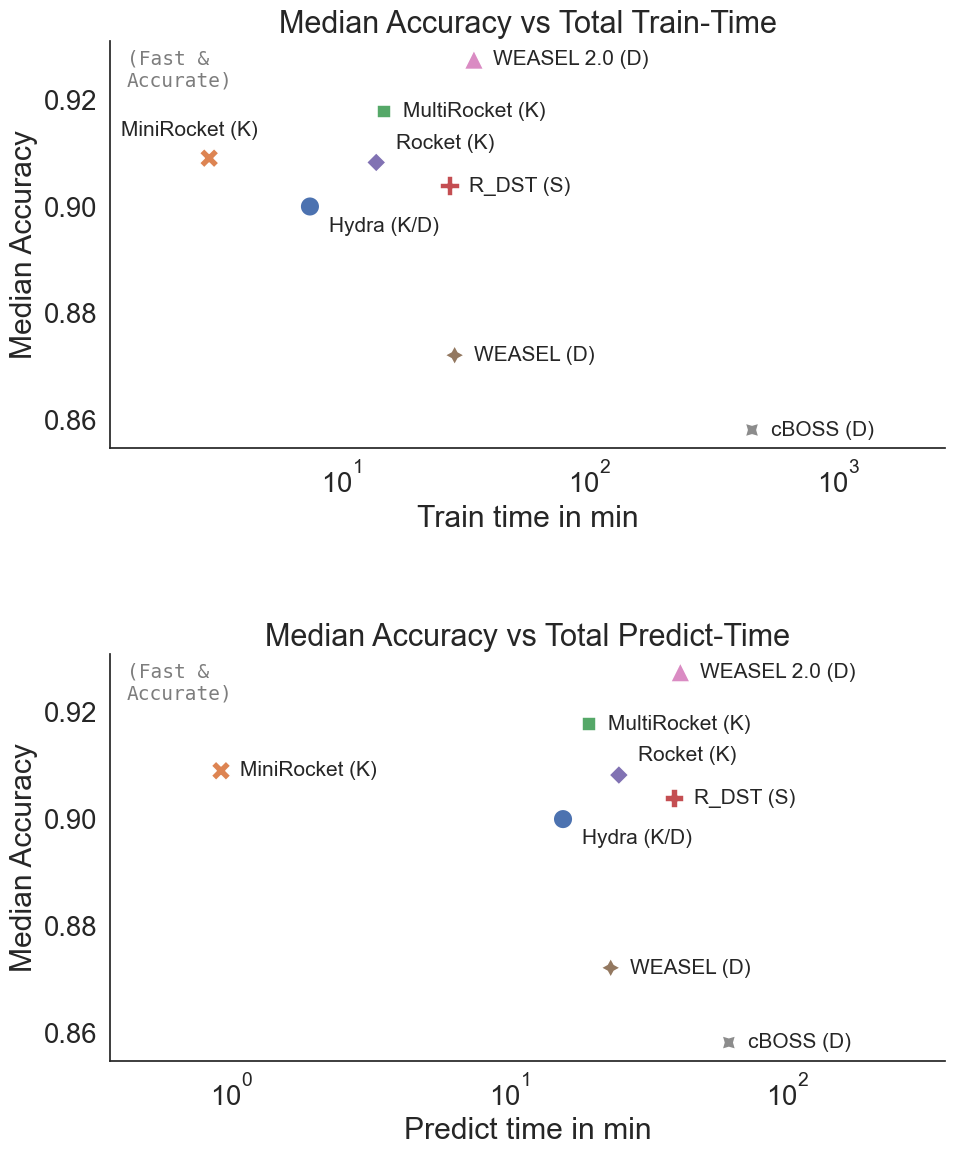}
\caption{Reproduced trade-off plots between median accuracy and total runtime across 114 UCR datasets. The upper-left corner is ideal (fast and accurate). WEASEL~2.0 achieves the highest median accuracy with moderate runtime.}
\label{fig:tradeoff}
\end{figure}

Figure~\ref{fig:runtime} shows the runtime distributions, and Figure~\ref{fig:tradeoff} shows the accuracy-runtime trade-off. MiniRocket is the fastest classifier (0.05\,h total fit), followed by Hydra (0.12\,h) and ROCKET (0.23\,h). WEASEL~2.0 requires 0.57\,h total fit time, which is moderate but substantially faster than cBOSS (7.51\,h). WEASEL~2.0 occupies the desirable upper-left region of the accuracy versus train-time scatter (high accuracy, moderate runtime), consistent with the paper's results. One discrepancy is that our reproduced total fit time (0.57\,h) and predict time (0.69\,h) are lower than the published values (1.71\,h and 1.79\,h, respectively). Our reproduction used aeon 1.3.0 and Python 3.12 on an AMD Ryzen AI 7 350, which differs from the original setup; we therefore do not interpret the speedup as a property of WEASEL~2.0 itself.

Overall, the paper's core claims are well supported by our reproduction. WEASEL~2.0 is the most accurate dictionary classifier, it is not statistically distinguishable from the best kernel methods, and it has a predictable memory footprint. The multi-comparison matrix (Figure~\ref{fig:mcm}, Appendix~\ref{apdx:mcm}) confirms that pairwise Wilcoxon tests show no significant difference between WEASEL~2.0 and four of the five top methods (R-DST, MiniRocket, Hydra, ROCKET), with $p$-values well above 0.05. MultiRocket is the exception, with WEASEL~2.0 significantly worse ($p = 0.029$). WEASEL~2.0 is significantly better than both WEASEL~1.0 and cBOSS ($p < 10^{-4}$).

At the per-dataset-type level, WEASEL~2.0 is strongest on EPG (0.992 mean), Hemodynamics (0.963), and ECG datasets (0.950), and weakest on EOG (0.544) and Device datasets (0.774). It wins or ties on the majority of Hemodynamics, Sensor, and Spectro datasets, consistent with the per-domain results in the original paper.

Several limitations of the reproduction are worth noting. The variable-length gesture datasets PickupGestureWiimoteZ ($+12.0\%$) and ShakeGestureWiimoteZ ($+8.0\%$) are clear outliers and the cause of this divergence could not be verified, as noted above. The reproduced critical difference diagram yields two overlapping cliques among the top six classifiers rather than the single clique shown in the original paper, which compared 16 classifiers; this is expected, as critical difference diagrams are sensitive to the addition or removal of methods. Finally, WEASEL~1.0 required reducing the number of threads to one for three datasets (Crop, ElectricDevices, FordB) due to out-of-memory errors. This same error was also seen for Hydra on the StarLightCurves dataset. These issues did not affect WEASEL~2.0 itself but slightly limit the soundness of the competitor comparison.

\clearpage
\section{Sensitivity Analysis of Design Choices}
\label{sec:sensitivity}

We investigate four design choices in WEASEL~2.0 that the original paper does not empirically justify:
\begin{enumerate}[label=(\alph*)]
    \item The choice of ridge as the downstream classifier.
    \item The absence of TF-IDF feature weighting.
    \item The maximum window-size heuristic.
    \item The maximum ensemble-size heuristic.
\end{enumerate}

For each, we test whether the original choice is empirically optimal. The first three prove robust to perturbation; the fourth is over-provisioned for long-series datasets, motivating the adaptive rule presented in Section~\ref{sec:adaptive}.

Preliminary experiments for (a)--(c) were conducted on a subset of 22 UCR datasets selected to span the full range of dataset characteristics: training set size (20 to 8926 instances), series length (24 to 2844 time points), and number of classes (2 to 50). The subset includes small-train short-series datasets (e.g.\ ItalyPowerDemand, SonyAIBORobotSurface1), small-train long-series datasets (e.g.\ HouseTwenty, Rock), large-train datasets (e.g.\ ElectricDevices, Crop), and many-class datasets (e.g.\ FiftyWords, Phoneme). This breadth ensures that results are not biased toward any single region of the dataset characteristic space. Definitive evaluation of the proposed variants in (d) is conducted on the full 112 fixed-length dataset benchmark to validate efficiency and accuracy trade-offs.

\subsection{Alternative Classifiers}
We replaced the default ridge classifier with a stochastic gradient descent (SGD) classifier to test whether a different optimisation strategy and regularisation scheme could improve accuracy, training time, or memory use. An SGD classifier was chosen because its cost grows linearly rather than quadratically with the number of training instances, which we hypothesised would reduce fit time on the larger UCR datasets. The experiment resulted in a statistically significant reduction in accuracy across all dataset types (mean $-0.069$, $p < 0.001$). Fit time increased significantly (mean $+8.19$\,s, $p < 0.001$), while predict time decreased (mean $-0.79$\,s, $p < 0.001$). Memory usage was essentially unchanged (mean $+0.55$\,MB, $p = 0.042$). The SGD classifier failed to match the generalisation performance of the ridge classifier, and results are summarised alongside the feature-weighting variants in Table~\ref{tab:feature_results}.

\subsection{Feature Weighting}
WEASEL~2.0 uses raw word counts as its feature values, applying no term weighting \cite{weasel2.0}. TF-IDF weighting has been used in earlier bag-of-words time series classifiers such as SAX-VSM~\cite{saxvsm}, where it is applied to downweight words common across all time series. We investigated whether applying TF-IDF to the WEASEL~2.0 word-count feature matrix could similarly improve discrimination. The variant was run on the subset of 22 UCR datasets. The results show a statistically significant reduction in accuracy (mean $-0.008$, $p = 0.015$) and a significant increase in fit time, while predict time decreased slightly. Memory usage increased by a mean of 43\,MB ($p = 0.038$).

Sublinear TF scaling was also evaluated to assess whether compressing high word counts improves accuracy or performance. Sublinear TF scaling replaces each raw term frequency $\mathrm{tf}$ with $1 + \log(\mathrm{tf})$~\cite{manning2008ir}, so that additional occurrences of a word contribute
progressively less to the feature value. The results show no statistically significant change in accuracy (mean $+0.002$, $p = 0.794$), suggesting the transformation is neutral to accuracy. However, fit time increased significantly and memory usage increased by a mean of 43\,MB ($p = 0.042$), while predict time decreased slightly. Neither variant offers a favourable trade-off relative to the baseline. Table~\ref{tab:feature_results} summarises the results for both variants alongside the SGD results.

\begin{table}[htb!]
\caption{Mean and median differences versus the WEASEL~2.0 baseline for the SGD, TF-IDF, and Sublinear TF variants on 22 UCR datasets. Bold indicates improvement over the baseline.}
\label{tab:feature_results}
\centering
\begin{tabular}{lrrrrrr}
\toprule
 & \multicolumn{2}{c}{SGD} & \multicolumn{2}{c}{TF-IDF} & \multicolumn{2}{c}{Sublinear TF} \\
\cmidrule(lr){2-3} \cmidrule(lr){4-5} \cmidrule(lr){6-7}
 & Mean & Median & Mean & Median & Mean & Median \\
\midrule
Fit time (s)     & $+8.19$ & $+1.33$ & $+40.74$ & $+1.69$ & $+42.90$ & $+1.68$ \\
Predict time (s) & $\mathbf{-0.79}$ & $\mathbf{-0.39}$ & $\mathbf{-0.20}$ & $\mathbf{-0.26}$ & $\mathbf{-0.16}$ & $\mathbf{-0.23}$ \\
Peak memory (MB) & $+0.55$ & $+0.07$ & $+43.20$ & $+0.07$ & $+42.72$ & $+0.06$ \\
Accuracy (\%)    & $-6.9$ & $-3.4$ & $-0.8$ & $-0.3$ & $\mathbf{+0.2}$ & $0.0$ \\
\bottomrule
\end{tabular}
\end{table}

\subsection{Maximum Window Size}
WEASEL~2.0 sets the maximum window size $w_{max}$ by the following rule:
\begin{equation}
w_{max} =
\begin{cases}
24 & \text{if } m < 250 \\
44 & \text{if } m \geq 250 \text{ and } n < 100 \\
84 & \text{else}
\end{cases}
\end{equation}
where $m$ is the number of training instances and $n$ is the series length. To assess whether these thresholds are well chosen, we swept seven values of $w_{max} \in \{12, 24, 44, 84, 100, 124, 200\}$ across the 22-dataset subset, holding ensemble size fixed at 50 to isolate the effect of window size alone. The set of tested values includes the three tiers of the original rule (24, 44, 84) and extends beyond them in both directions to test whether smaller or larger windows offer any benefit to particular datasets.

For datasets with series length shorter than the requested maximum window (e.g.\ ItalyPowerDemand, $n = 24$), the value is capped to the length of the series. These datasets produced identical results across any larger window values. This is expected behaviour and does not affect the validity of the comparison.

A Wilcoxon signed-rank test across the 22 datasets shows no statistically significant difference in accuracy between any tested window value and the baseline WEASEL~2.0 results ($p > 0.05$), with the exception of $w_{max} = 100$ and $w_{max} = 124$, which are significantly worse. This indicates that the original rule of thumb is robust and well chosen. Increasing the maximum window beyond the current ceiling of 84 provides no accuracy benefit and can be harmful in some cases. We therefore retain the original maximum window rule unchanged in our proposed variant.

\subsection{Maximum Ensemble Size}
WEASEL~2.0 sets the maximum ensemble size $r_{max}$ based on the training set size $m$ and series length $n$. Like $w_{max}$, this rule is presented in the original paper as a simple rule of thumb without empirical justification for the chosen thresholds:
\begin{equation}
r_{max} =
\begin{cases}
50  & \text{if } m < 250 \\
100 & \text{if } m \geq 250 \text{ and } n < 100 \\
150 & \text{else}
\end{cases}
\end{equation}
This creates sharp transitions: a dataset with 249 instances receives 50 ensemble members (approx.\ 25k features), while one with 250 receives 100 (approx.\ 51k features). Of the 114 UCR datasets, 62 (54\%) trigger the small-dataset branch, 14 (12\%) trigger the short-series branch, and 38 (33\%) use the default.

To characterise the relationship between ensemble size and accuracy, we tested a range of $r_{max} \in \{10, 20, 25, 30, 50, 75, 100, 125, 150, 175\}$ on the 22-dataset subset. Very small ensembles ($r_{max} \leq 20$) degrade accuracy noticeably, establishing 25 as a practical lower bound. Closer analysis of the larger UCR datasets showed that the original rule is wasteful for large datasets: for $n > 500$, each additional 25 ensemble members cost hundreds to thousands of MB of memory for marginal or negative accuracy returns, with the optimal $r_{max}$ for most large datasets lying between 50 and 75 rather than the default 150. This diagnostic finding, that the ensemble-size heuristic is over-provisioned for long-series datasets, motivates the adaptive rule presented in Section~\ref{sec:adaptive}.

\section{An Adaptive Ensemble-Size Rule}
\label{sec:adaptive}

The diagnostic finding of Section~\ref{sec:sensitivity}, that the ensemble-size heuristic is over-provisioned for long-series datasets, suggests that the default $r_{max}$ values can be reduced for some dataset regimes without accuracy loss. We first tested a simple uniform reduction. We evaluated halving all three tiers of the original rule, from $50/100/150$ to $25/50/75$, on 112 fixed-length UCR datasets. This produced statistically significant reductions in fit time (mean $-3.6$\,s, median $-0.9$\,s), predict time (mean $-4.1$\,s, median $-0.7$\,s), and peak memory (mean $-306$\,MB, median $-52$\,MB). The disparity between mean and median reflects that savings are concentrated on large datasets where the original ensemble was largest. However, halving also produced a statistically significant reduction in mean accuracy of $0.22\%$, which motivated a more targeted approach.

Rather than uniformly reducing ensemble size, we propose an adaptive rule that selectively preserves larger ensembles for problem types identified as sensitive and reduces them elsewhere. Empirical analysis of the fixed $r_{max}$ sweep showed that series length and number of classes are the primary drivers of ensemble sensitivity, while training set size was not a reliable predictor and was excluded from the rule. Long series require more configurations to adequately sample the larger SFA vocabulary space, while problems with fewer classes require fewer features to discriminate between them. This leads to the following rule:
\begin{equation}
r_{max} =
\begin{cases}
75 & \text{if } n > 700 \\
25 & \text{if } c \leq 2 \\
50 & \text{else}
\end{cases}
\label{eq:adaptive}
\end{equation}
where $n$ is the series length and $c$ is the number of classes; where both conditions apply, series length takes priority. The series-length threshold $n > 700$ is the smallest value at which our sweep showed accuracy gains from $r_{max} > 75$ across multiple long-series datasets. The binary-class branch ($c \leq 2 \to 25$) reflects that binary problems achieve baseline accuracy with smaller feature counts. The ensemble values $25, 50, 75$ are $\sfrac{1}{2}$, $1$, and $\sfrac{3}{2}$ of the default $50$, chosen to keep the rule interpretable. Relative to the halved-ensemble variant, this rule increases $r_{max}$ for long-series datasets while reducing it more aggressively for binary and short-series problems.

\begin{figure*}[htb!]
\centering
\begin{subfigure}[b]{0.32\textwidth}
\centering
\includegraphics[width=\textwidth]{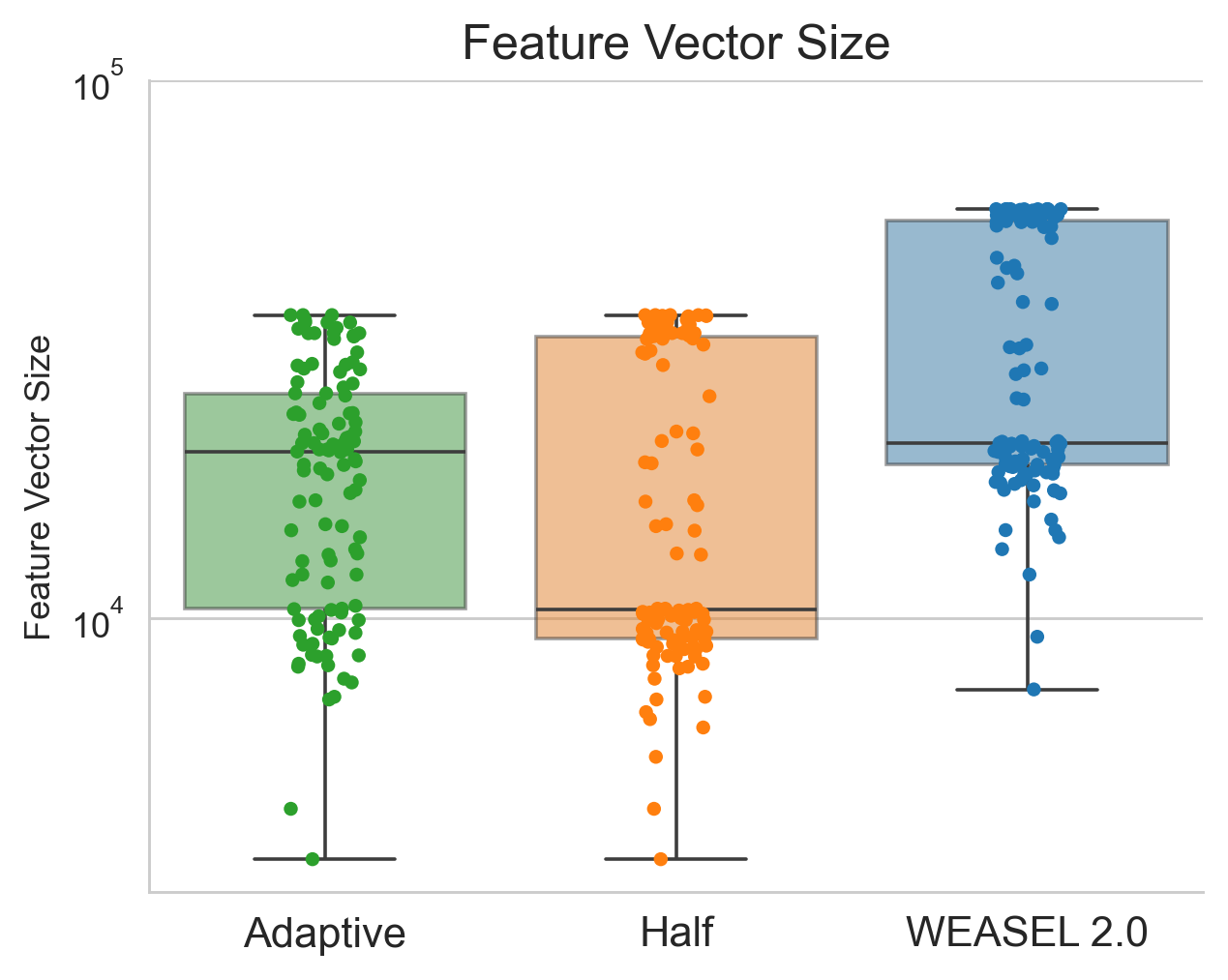}
\caption{Feature vector size.}
\label{fig:features_res}
\end{subfigure}
\hfill
\begin{subfigure}[b]{0.32\textwidth}
\centering
\includegraphics[width=\textwidth]{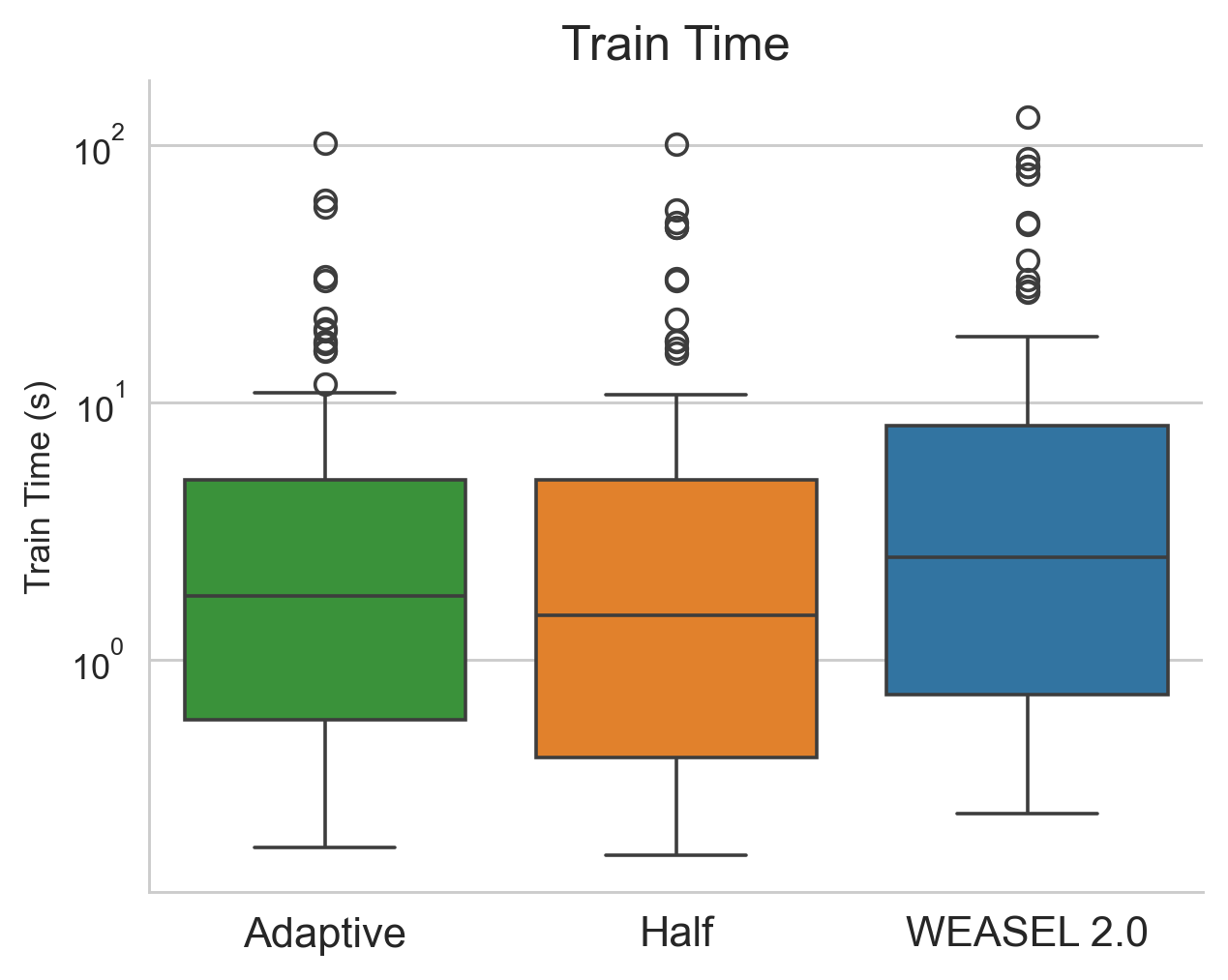}
\caption{Training time.}
\label{fig:train-time}
\end{subfigure}
\hfill
\begin{subfigure}[b]{0.32\textwidth}
\centering
\includegraphics[width=\textwidth]{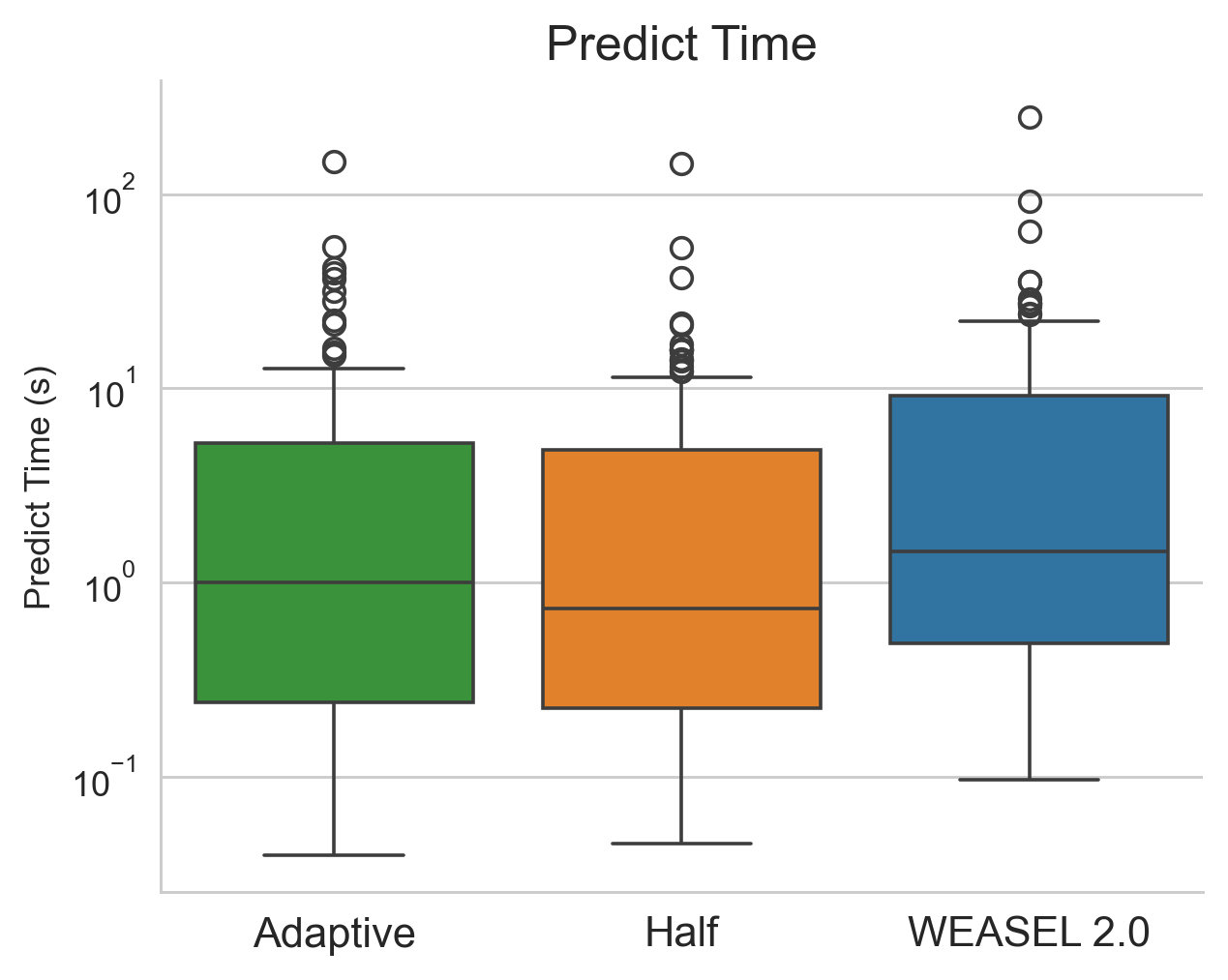}
\caption{Prediction time.}
\label{fig:predict-time}
\end{subfigure}

\vspace{6pt}

\begin{subfigure}[b]{0.48\textwidth}
\centering
\includegraphics[width=\textwidth]{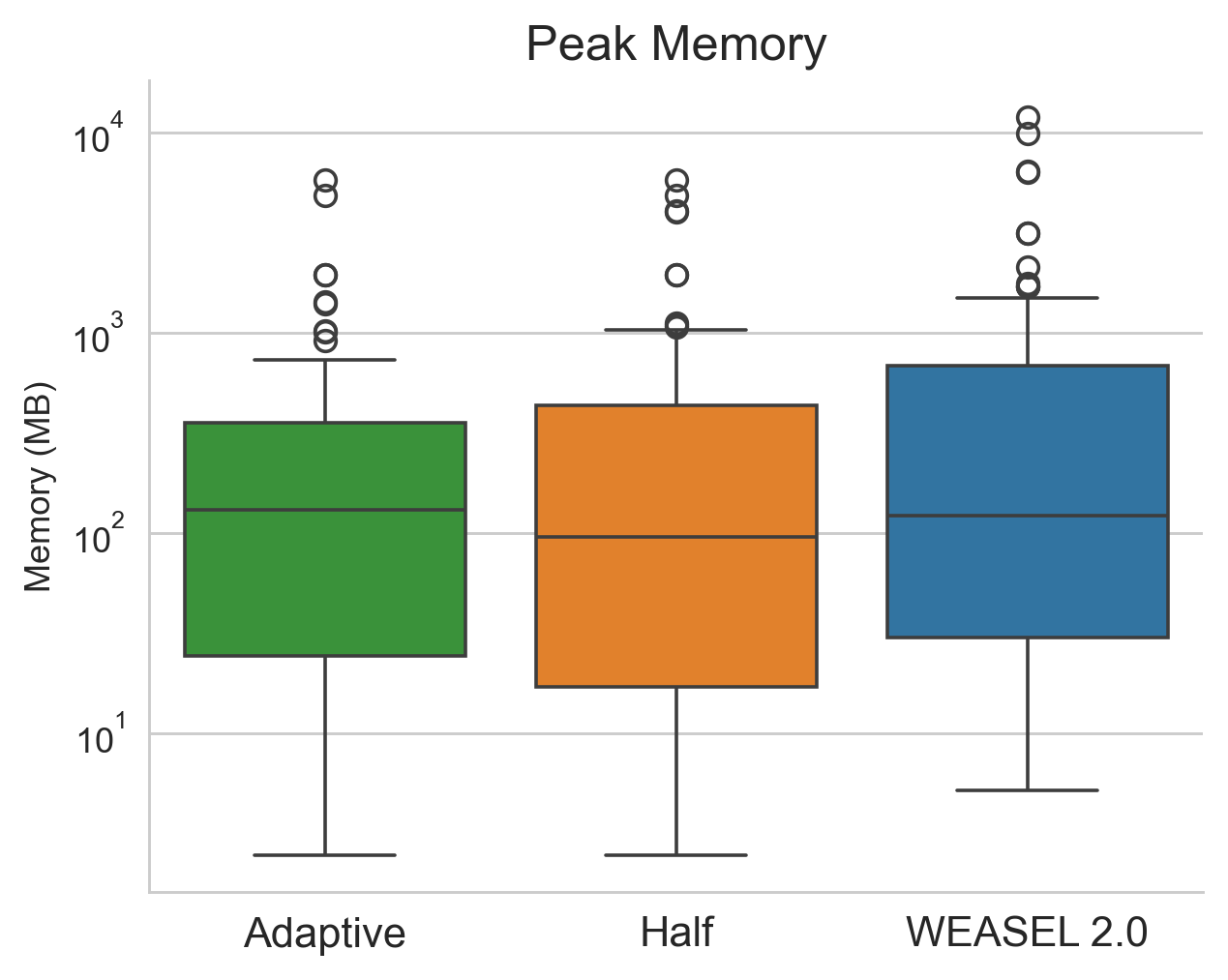}
\caption{Peak fit memory.}
\label{fig:memory}
\end{subfigure}
\hfill
\begin{subfigure}[b]{0.48\textwidth}
\centering
\includegraphics[width=\textwidth]{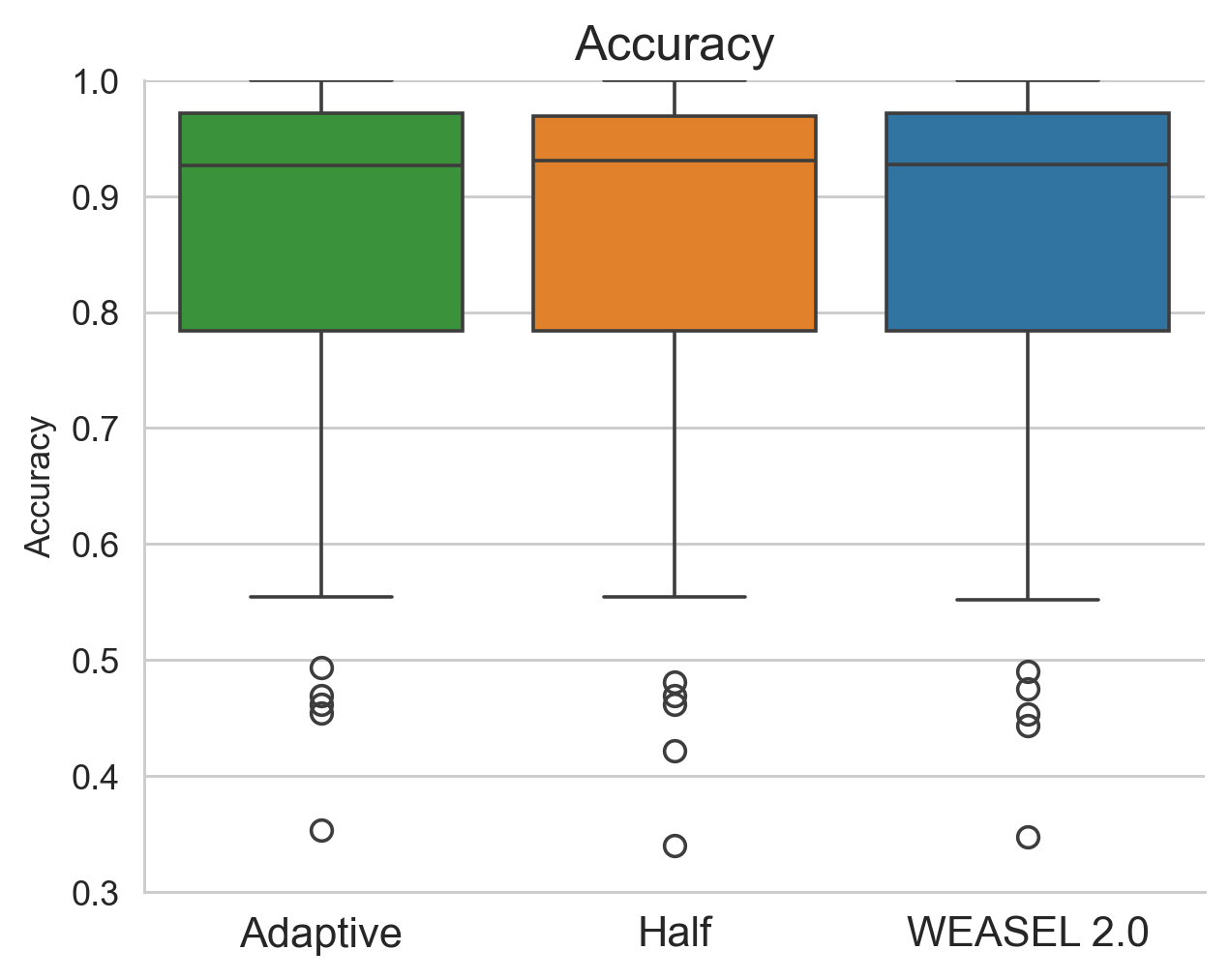}
\caption{Median accuracy.}
\label{fig:accuracy}
\end{subfigure}
\caption{Performance comparison for the new variants versus the WEASEL~2.0 baseline across five metrics on the UCR benchmark.}
\label{fig:results}
\end{figure*}

Evaluated on 112 fixed-length datasets, the adaptive rule achieves statistically significant reductions in fit time (mean $-4.0$\,s, median $-0.4$\,s), predict time (mean $-3.4$\,s, median $-0.3$\,s), and peak memory (mean $-395$\,MB, median $-37$\,MB). The mean accuracy reduction is $0.11\%$ with a median of $0.00\%$, indicating that more than half of datasets show no accuracy change. Accuracy is fully preserved across small training sets, short and medium series, and binary classification tasks. A modest accuracy cost is observed for large training sets and many-class problems. One notable exception is PigAirwayPressure, where the extreme class-to-sample ratio means a larger ensemble remains advisable; this dataset presents a challenge for the original WEASEL~2.0 as well. Table~\ref{tab:ensemble_results} compares the halved and adaptive variants directly against the baseline, and Figure~\ref{fig:results} shows the full distribution of feature counts, runtimes, memory, and accuracy across the three variants.

\begin{figure}[ht!]
\centering
\includegraphics[width=.8\textwidth]{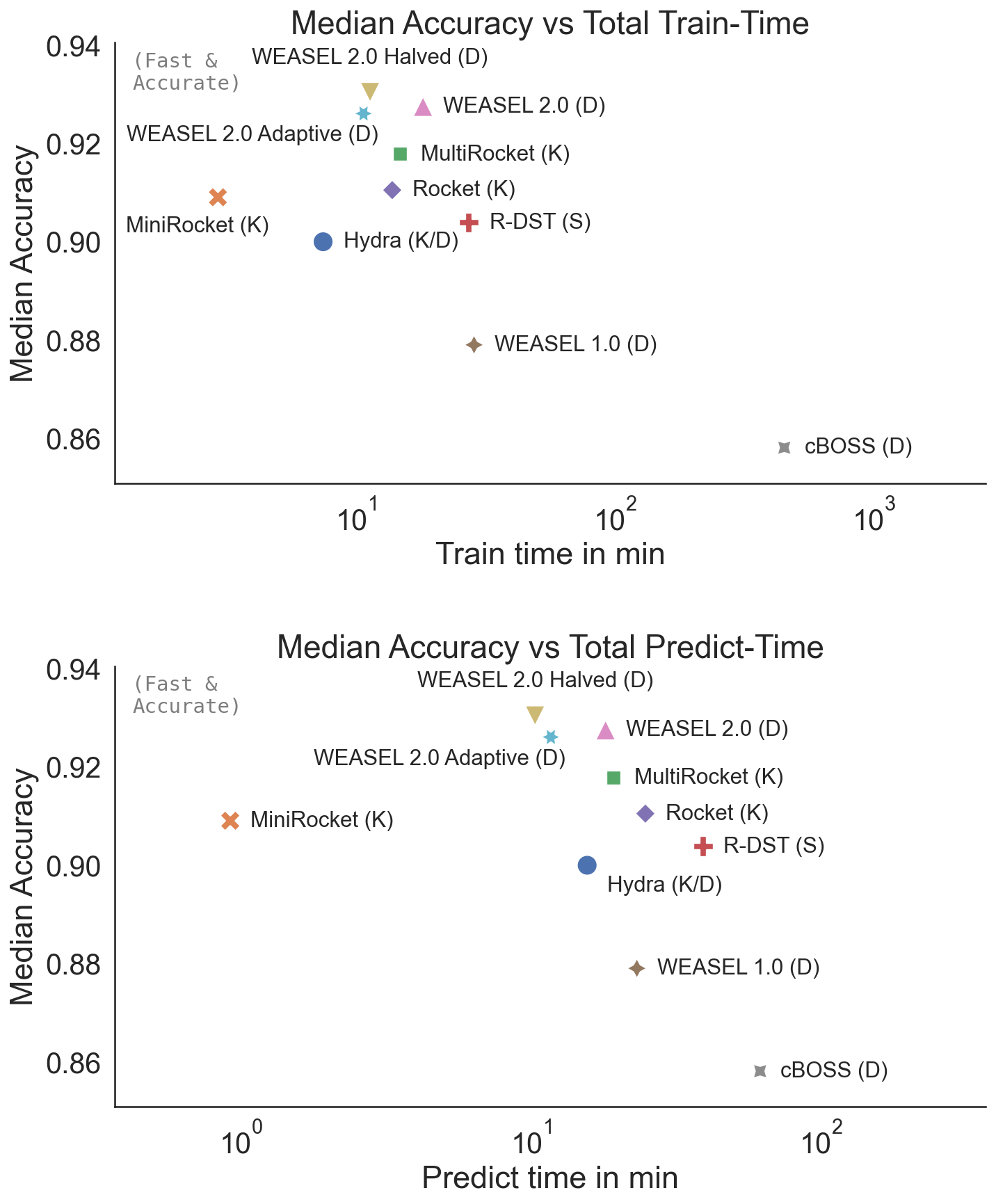}
\caption{Trade-off plots between median accuracy and total runtime across 112 UCR datasets. The upper-left corner is ideal (fast and accurate). Comparison between WEASEL~2.0, adaptive, and halved variants.}
\label{fig:tradeoff_v2}
\end{figure}

\begin{figure}[htbp]
\centering
\includegraphics[width=\columnwidth]{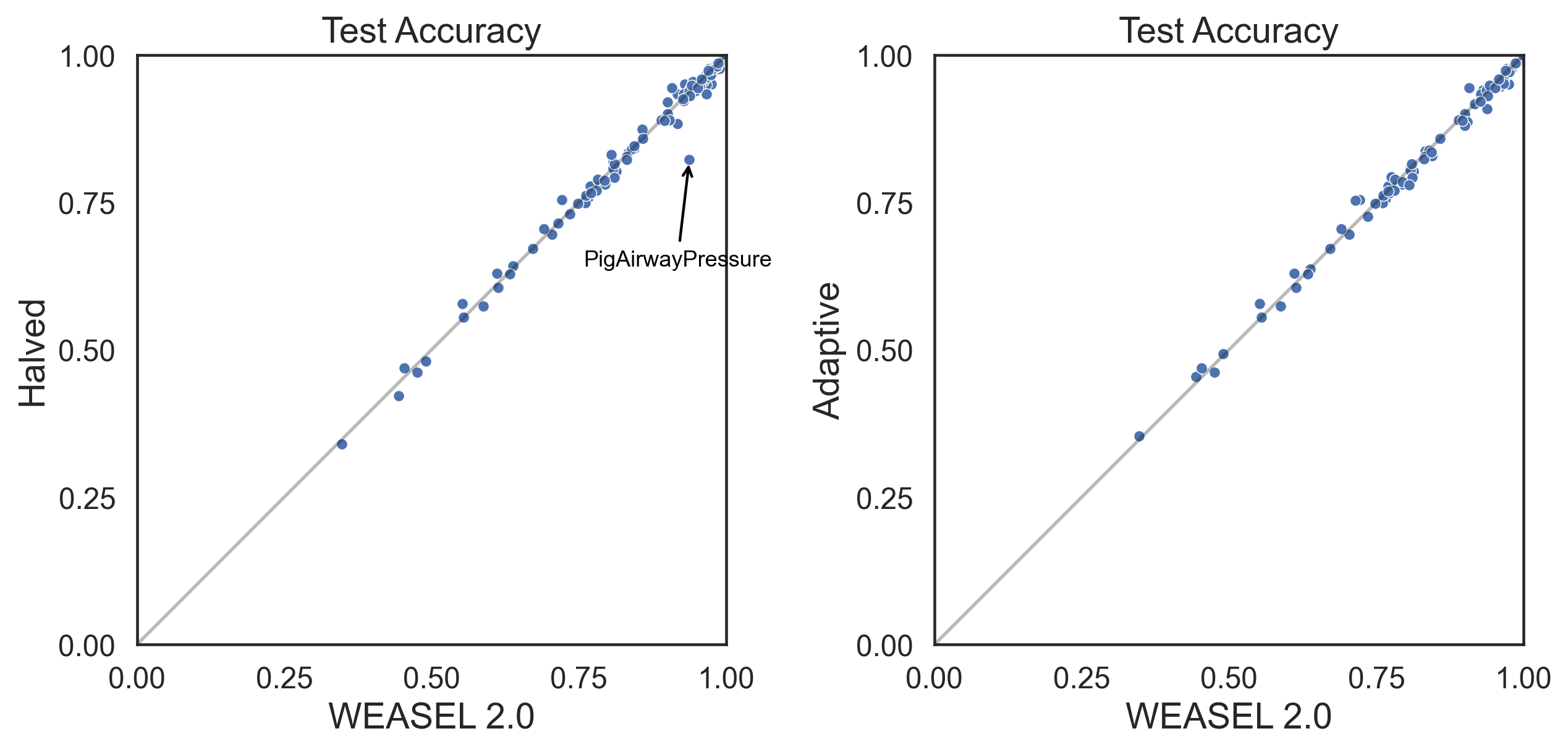}
\caption{Pairwise comparison of the adaptive and halved variants against WEASEL~2.0. A point below the diagonal indicates that WEASEL~2.0 is more accurate than the variant on that dataset.}
\label{fig:scatter}
\end{figure}

\begin{figure}[htb!]
\centering
\includegraphics[width=\columnwidth]{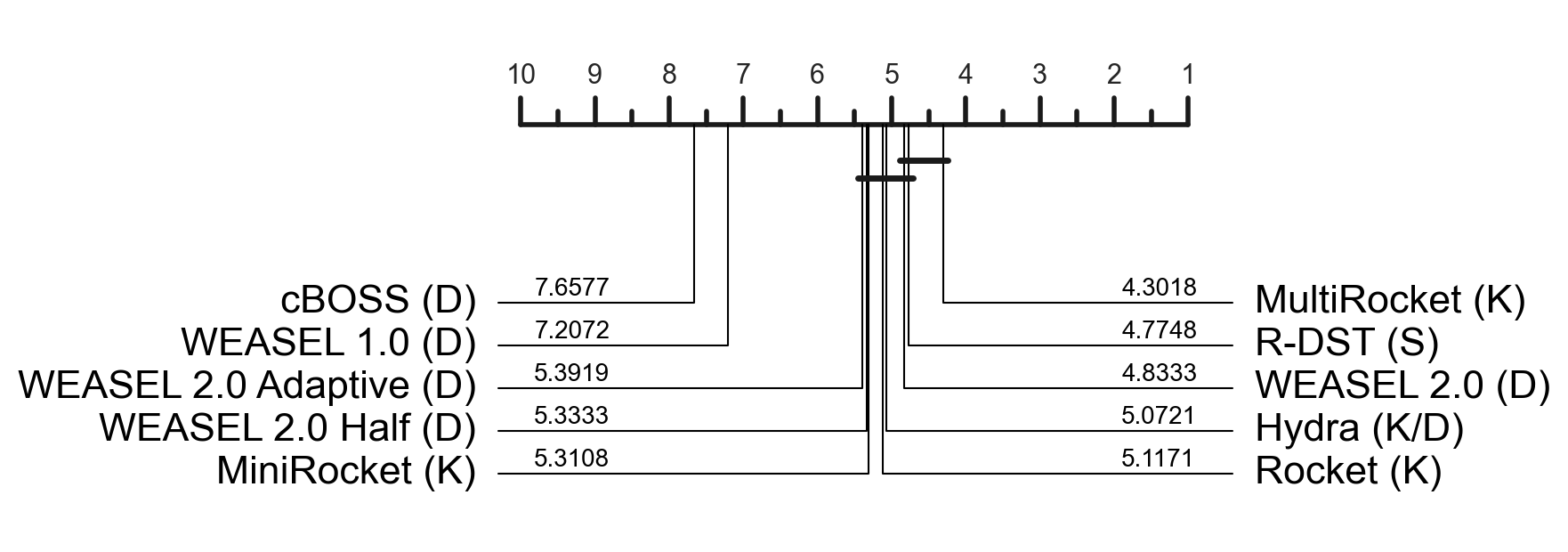}
\caption{Critical difference diagram on mean ranks across the UCR datasets. Methods connected by a horizontal bar are not significantly different under the Nemenyi test at $\alpha = 0.05$.}
\label{fig:cd_all}
\end{figure}

\begin{table}[ht!]
\caption{Mean and median differences versus the WEASEL~2.0 baseline for the halved-ensemble and adaptive-ensemble variants. Bold indicates the better of the two variants in each row. All differences are statistically significant ($p < 0.05$).}
\label{tab:ensemble_results}
\centering
\begin{tabular}{lrrrr}
\toprule
 & \multicolumn{2}{c}{Halved} & \multicolumn{2}{c}{Adaptive} \\
\cmidrule(lr){2-3} \cmidrule(lr){4-5}
 & Mean & Median & Mean & Median \\
\midrule
Fit time (s)     & $-3.64$ & $\mathbf{-0.88}$ & $\mathbf{-3.99}$ & $-0.43$ \\
Predict time (s) & $\mathbf{-4.15}$ & $\mathbf{-0.71}$ & $-3.41$ & $-0.32$ \\
Peak memory (MB) & $-306.4$ & $\mathbf{-51.9}$ & $\mathbf{-394.7}$ & $-37.4$ \\
Accuracy (\%)    & $-0.22$ & $0.00$  & $\mathbf{-0.11}$ & $0.00$ \\
\bottomrule
\end{tabular}
\end{table}
Figure~\ref{fig:tradeoff_v2} places the two variants on the accuracy-runtime trade-off plane against the WEASEL~2.0 baseline; both variants move into the upper-left region (faster, comparable accuracy) without leaving the cluster of competitive methods. Figure~\ref{fig:scatter} shows pairwise accuracy comparisons between WEASEL~2.0 and each variant across 112 datasets. The vast majority of points lie on or close to the diagonal for both variants, confirming that accuracy is largely preserved across the benchmark. The adaptive variant shows tighter clustering around the diagonal than the halved variant. The one notable outlier below the diagonal in the halved-ensemble variant is PigAirwayPressure, where the reduced ensemble size is insufficient; the adaptive rule retains the larger ensemble for this dataset and recovers the accuracy. Figure~\ref{fig:cd_all} shows the critical difference diagram across all 10 classifiers including the two variants; both are statistically indistinguishable from the original WEASEL~2.0 under the Nemenyi test.

The adaptive rule is preferable to simple halving. It achieves greater mean memory savings (395\,MB vs.\ 306\,MB) and roughly half the accuracy cost ($-0.11\%$ vs.\ $-0.22\%$), while the halved-ensemble variant produces slightly larger median memory and time savings on smaller datasets. The feature vector size distribution confirms that both variants shift feature counts downward relative to the baseline while remaining well within the controlled range that distinguishes WEASEL~2.0 from the unbounded feature spaces of WEASEL~1.0. The practical value of these results is most evident in settings where many runs accumulate, such as large-scale benchmark sweeps or repeated multi-seed experiments, where memory and time savings of a few hundred megabytes and seconds per run compound into a meaningful reduction in total compute budget.

\section{Conclusion}
\label{sec:conclusion}

We have presented an independent reproduction and sensitivity analysis of \linebreak WEASEL~2.0. Our reproduction on 114 UCR datasets matches the published values within statistical noise, supporting the original paper's three main claims: WEASEL~2.0 is the strongest dictionary classifier in the benchmark, it is not statistically distinguishable from the best kernel and shapelet methods, and it has a controlled memory footprint.

Our sensitivity analysis of four under-justified design choices shows that three of them, the ridge classifier, the absence of feature weighting, and the maximum window-size heuristic, are empirically robust. Replacing ridge with SGD or applying TF-IDF weighting degraded accuracy; sublinear TF scaling was neutral to accuracy but increased fit time and memory. Varying the maximum window size around the original tiers produced no significant accuracy gain. The fourth design choice, the maximum ensemble-size heuristic, was over-provisioned for long-series datasets, where additional ensemble members cost substantial memory for marginal accuracy returns.

We proposed an adaptive ensemble-size rule that selects $r_{max}$ based on series length and number of classes, rather than the original criterion of training set size. Evaluated on 112 fixed-length datasets, the adaptive rule reduces peak fit memory by a median of 37\,MB (mean 395\,MB) and fit time by a median of 0.4\,s (mean 4\,s), with a median accuracy change of zero. The savings concentrate on long-series datasets where the original rule allocated the largest ensembles, while accuracy is preserved on small training sets, short and medium series, and binary classification tasks.

Two natural extensions of this work remain open. The first is a data-driven version of the rule that selects $r_{max}$ from cross-validated accuracy during fit, rather than from static thresholds; this would adapt to dataset properties not captured by series length and class count alone. The second is
the evaluation of the adaptive rule in a multivariate setting. Memory demand grows with the number of channels, so the savings of the adaptive rule could be greater than in the univariate case. The rule would transfer naturally if per-channel series length is used as input to the threshold.

\begin{credits}
\subsubsection{\ackname} This work was partly funded by Taighde Éireann -- Research Ireland through the Research Ireland Centre for Research Training in Machine Learning (18/CRT/6183). We would also like to thank the UCD School of Computer Science for funding our conference registration and attendance. The authors used Claude (Anthropic) to assist with language editing and with drafting and debugging experiment scripts. All code was reviewed and executed by the authors, and all reported results were produced by that code. The authors take full responsibility for all content in this paper.

\subsubsection{\discintname}
The authors have no competing interests to declare that are relevant to the content of this article.
\end{credits}

\newpage

\bibliographystyle{splncs04}
\bibliography{references}

\newpage

\appendix
\section{Reproduction Results}
\begin{longtable}{lrrrr}
\caption{WEASEL~2.0 reproduction per-dataset results on 114 UCR datasets.}\\
\label{tab:full_results}\\
\toprule
Dataset & Accuracy & \makecell{Fit\\time (s)} & \makecell{Predict\\time (s)} & \makecell{Peak\\memory (MB)} \\
\midrule
\endfirsthead
\toprule
Dataset & Accuracy & \makecell{Fit\\time (s)} & \makecell{Predict\\time (s)} & \makecell{Peak\\memory (MB)} \\
\midrule
\endhead
\midrule
\multicolumn{5}{r}{\footnotesize Continued on next page} \\
\endfoot
\bottomrule
\endlastfoot
ACSF1 & 0.940 & 4.08 & 3.02 & 90.6 \\
Adiac & 0.841 & 11.66 & 5.05 & 558.8 \\
ArrowHead & 0.857 & 0.75 & 1.25 & 21.6 \\
BME & 0.980 & 0.51 & 0.63 & 18.6 \\
Beef & 0.833 & 0.97 & 0.42 & 19.0 \\
BeetleFly & 0.950 & 0.71 & 0.29 & 12.8 \\
BirdChicken & 0.900 & 0.71 & 0.29 & 12.6 \\
CBF & 0.982 & 0.55 & 3.39 & 20.1 \\
Car & 0.917 & 1.57 & 0.91 & 47.1 \\
Chinatown & 0.974 & 0.34 & 0.35 & 5.2 \\
ChlorineConcentration & 0.767 & 12.30 & 57.84 & 778.5 \\
CinCECGTorso & 0.962 & 2.54 & 57.41 & 29.6 \\
Coffee & 1.000 & 0.79 & 0.27 & 14.9 \\
Computers & 0.700 & 22.28 & 15.89 & 434.5 \\
CricketX & 0.813 & 15.63 & 9.15 & 687.6 \\
CricketY & 0.808 & 15.49 & 9.27 & 687.3 \\
CricketZ & 0.808 & 16.52 & 9.61 & 687.5 \\
Crop & 0.761 & 117.91 & 61.07 & 9836.2 \\
DiatomSizeReduction & 0.967 & 0.70 & 2.79 & 6.7 \\
DistalPhalanxOutlineAgeGroup & 0.770 & 5.49 & 0.68 & 390.2 \\
DistalPhalanxOutlineCorrect & 0.779 & 7.28 & 1.44 & 703.6 \\
DistalPhalanxTW & 0.698 & 4.89 & 0.58 & 394.3 \\
ECG200 & 0.890 & 0.98 & 0.33 & 62.8 \\
ECG5000 & 0.948 & 11.59 & 39.99 & 888.5 \\
ECGFiveDays & 0.970 & 0.38 & 2.78 & 14.0 \\
EOGHorizontalSignal & 0.616 & 36.20 & 29.50 & 631.6 \\
EOGVerticalSignal & 0.472 & 38.25 & 29.70 & 632.6 \\
Earthquakes & 0.748 & 17.26 & 5.31 & 545.6 \\
ElectricDevices & 0.773 & 181.80 & 37.02 & 11925.0 \\
EthanolLevel & 0.588 & 73.41 & 57.86 & 821.3 \\
FaceAll & 0.786 & 9.37 & 12.53 & 982.9 \\
FaceFour & 0.943 & 0.61 & 0.77 & 16.3 \\
FacesUCR & 0.947 & 1.34 & 5.71 & 130.8 \\
FiftyWords & 0.833 & 13.50 & 7.34 & 736.5 \\
Fish & 0.989 & 2.80 & 1.71 & 99.8 \\
FordA & 0.961 & 157.42 & 39.60 & 6327.1 \\
FordB & 0.840 & 147.25 & 22.14 & 6390.5 \\
FreezerRegularTrain & 0.993 & 1.84 & 16.96 & 95.3 \\
FreezerSmallTrain & 0.974 & 0.47 & 17.10 & 17.8 \\
GunPoint & 1.000 & 0.50 & 0.51 & 27.9 \\
GunPointAgeSpan & 1.000 & 1.04 & 0.98 & 71.9 \\
GunPointMaleVersusFemale & 0.997 & 1.10 & 1.07 & 70.9 \\
GunPointOldVersusYoung & 0.994 & 1.01 & 0.95 & 72.2 \\
Ham & 0.762 & 1.66 & 0.96 & 66.9 \\
HandOutlines & 0.959 & 186.94 & 57.01 & 1713.6 \\
Haptics & 0.490 & 4.52 & 6.06 & 101.6 \\
Herring & 0.672 & 1.18 & 0.67 & 37.2 \\
HouseTwenty & 0.966 & 2.23 & 4.61 & 34.8 \\
InlineSkate & 0.444 & 4.72 & 18.51 & 84.8 \\
InsectEPGRegularTrain & 1.000 & 1.43 & 2.93 & 40.9 \\
InsectEPGSmallTrain & 0.984 & 0.49 & 2.89 & 11.9 \\
InsectWingbeatSound & 0.637 & 1.96 & 9.23 & 130.1 \\
ItalyPowerDemand & 0.957 & 0.42 & 0.75 & 30.4 \\
LargeKitchenAppliances & 0.832 & 23.85 & 17.88 & 645.6 \\
Lightning2 & 0.738 & 1.34 & 0.80 & 36.5 \\
Lightning7 & 0.822 & 1.04 & 0.56 & 42.0 \\
Mallat & 0.938 & 1.71 & 42.92 & 32.5 \\
Meat & 0.917 & 1.14 & 0.60 & 27.6 \\
MedicalImages & 0.779 & 4.00 & 3.07 & 489.9 \\
MiddlePhalanxOutlineAgeGroup & 0.617 & 3.71 & 0.51 & 353.6 \\
MiddlePhalanxOutlineCorrect & 0.832 & 5.20 & 1.02 & 581.9 \\
MiddlePhalanxTW & 0.558 & 3.57 & 0.53 & 355.5 \\
MixedShapesRegularTrain & 0.975 & 37.50 & 137.21 & 859.5 \\
MixedShapesSmallTrain & 0.962 & 2.92 & 42.89 & 60.9 \\
MoteStrain & 0.961 & 0.37 & 2.37 & 11.5 \\
NonInvasiveFetalECGThorax1 & 0.927 & 102.77 & 84.42 & 3154.4 \\
NonInvasiveFetalECGThorax2 & 0.955 & 102.97 & 89.15 & 3154.7 \\
OSULeaf & 0.967 & 2.68 & 1.87 & 126.6 \\
OliveOil & 0.967 & 0.77 & 0.41 & 13.3 \\
PhalangesOutlinesCorrect & 0.839 & 16.62 & 2.83 & 2126.2 \\
Phoneme & 0.347 & 5.86 & 35.65 & 127.4 \\
PickupGestureWiimoteZ & 0.900 & 1.02 & 0.51 & 33.4 \\
PigAirwayPressure & 0.938 & 5.79 & 8.06 & 92.2 \\
PigArtPressure & 0.986 & 5.18 & 7.46 & 101.1 \\
PigCVP & 0.966 & 5.18 & 7.68 & 102.8 \\
Plane & 1.000 & 0.82 & 0.34 & 61.0 \\
PowerCons & 0.928 & 1.36 & 0.62 & 112.5 \\
ProximalPhalanxOutlineAgeGroup & 0.863 & 3.78 & 0.75 & 313.9 \\
ProximalPhalanxOutlineCorrect & 0.900 & 5.22 & 1.06 & 521.4 \\
ProximalPhalanxTW & 0.810 & 3.66 & 0.76 & 312.1 \\
RefrigerationDevices & 0.557 & 22.67 & 16.42 & 638.3 \\
Rock & 0.900 & 1.60 & 2.69 & 25.1 \\
ScreenType & 0.496 & 23.65 & 17.47 & 647.0 \\
SemgHandGenderCh2 & 0.938 & 37.41 & 60.58 & 532.1 \\
SemgHandMovementCh2 & 0.631 & 55.68 & 45.43 & 796.4 \\
SemgHandSubjectCh2 & 0.896 & 56.68 & 45.04 & 795.6 \\
ShakeGestureWiimoteZ & 0.960 & 0.83 & 0.41 & 64.5 \\
ShapeletSim & 1.000 & 0.51 & 1.74 & 13.5 \\
ShapesAll & 0.935 & 24.88 & 16.75 & 1039.2 \\
SmallKitchenAppliances & 0.776 & 24.46 & 18.23 & 644.9 \\
SmoothSubspace & 0.973 & 0.65 & 0.14 & 65.2 \\
SonyAIBORobotSurface1 & 0.940 & 0.32 & 0.96 & 11.4 \\
SonyAIBORobotSurface2 & 0.951 & 0.33 & 1.45 & 16.4 \\
StarLightCurves & 0.983 & 75.78 & 519.73 & 1701.1 \\
Strawberry & 0.981 & 14.88 & 5.22 & 999.2 \\
SwedishLeaf & 0.976 & 8.03 & 4.30 & 872.2 \\
Symbols & 0.984 & 0.47 & 7.41 & 14.1 \\
SyntheticControl & 0.997 & 2.64 & 0.79 & 414.9 \\
ToeSegmentation1 & 0.969 & 0.56 & 1.20 & 26.7 \\
ToeSegmentation2 & 0.938 & 0.55 & 0.89 & 23.9 \\
Trace & 1.000 & 1.16 & 0.57 & 63.6 \\
TwoLeadECG & 0.999 & 0.34 & 1.99 & 11.3 \\
TwoPatterns & 0.998 & 15.60 & 28.74 & 1758.0 \\
UMD & 0.986 & 0.42 & 0.48 & 23.2 \\
UWaveGestureLibraryAll & 0.958 & 62.05 & 186.96 & 1494.4 \\
UWaveGestureLibraryX & 0.843 & 24.79 & 60.26 & 1469.2 \\
UWaveGestureLibraryY & 0.769 & 24.35 & 61.08 & 1497.5 \\
UWaveGestureLibraryZ & 0.791 & 24.20 & 60.54 & 1491.2 \\
Wafer & 1.000 & 18.88 & 60.25 & 1709.2 \\
Wine & 0.907 & 0.66 & 0.25 & 16.7 \\
WordSynonyms & 0.735 & 7.05 & 9.01 & 454.6 \\
Worms & 0.714 & 4.40 & 1.34 & 104.5 \\
WormsTwoClass & 0.805 & 4.27 & 1.26 & 117.5 \\
Yoga & 0.927 & 10.92 & 67.77 & 522.4 \\
\end{longtable}

\newpage
\section{Multi-Comparison Matrix}
\label{apdx:mcm}
\begin{figure}[H]
\centering
\rotatebox{90}{%
    \begin{minipage}{\textheight}
        \includegraphics[width=\linewidth]{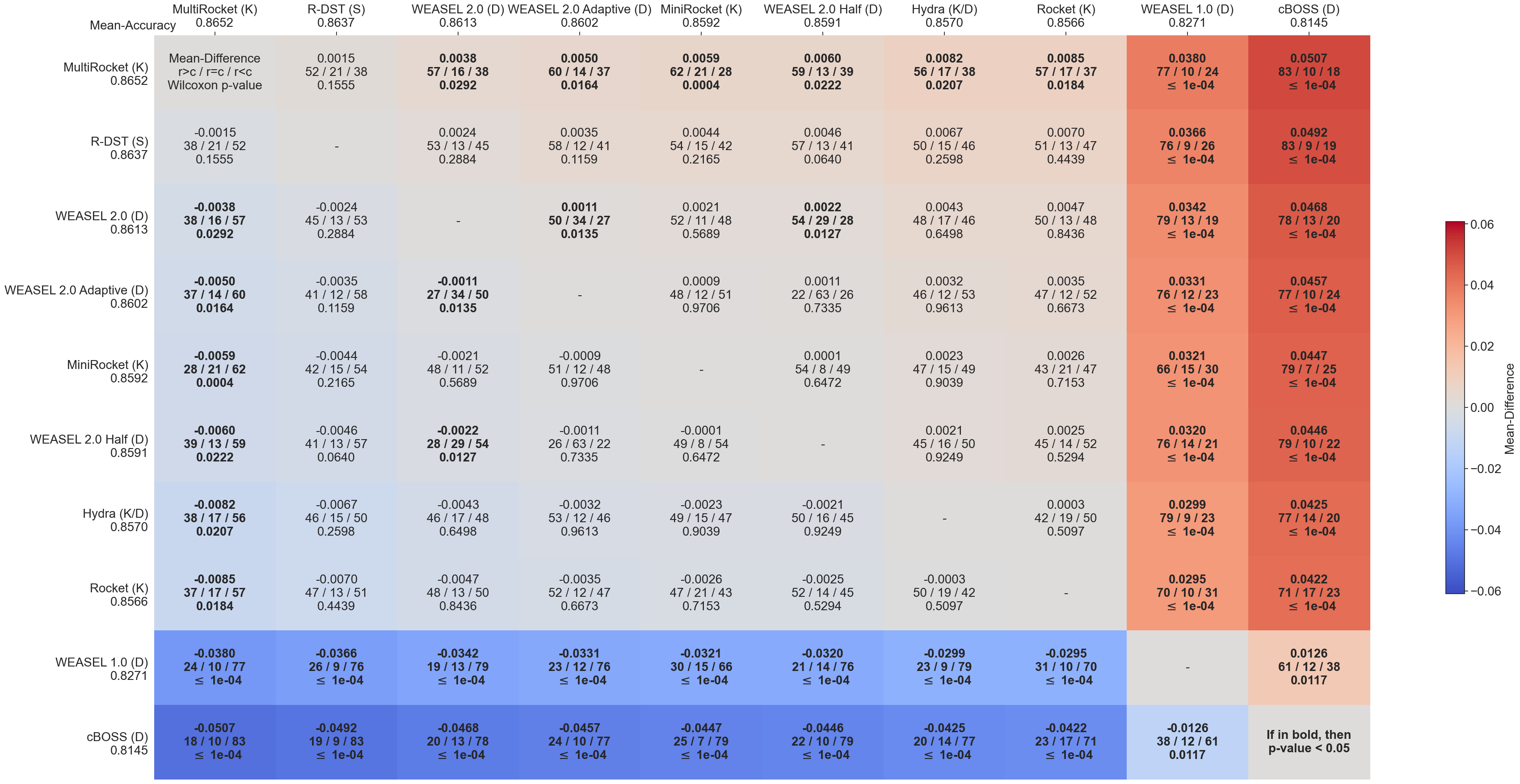}
        \captionof{figure}{Multi-comparison matrix of our reproduced results showing pairwise mean differences, win/tie/loss counts, and Wilcoxon $p$-values. Bold entries indicate $p < 0.05$.}
        \label{fig:mcm}
    \end{minipage}%
}
\end{figure}

\newpage
\section{Adaptive Ensemble-Size Rule Results}
\begin{longtable}{lrrrr}
\caption{WEASEL~2.0 new adaptive ensemble-size rule variant per-dataset results on fixed-length UCR datasets.}\\
\label{tab:ae_results}\\
\toprule
Dataset & Accuracy & \makecell{Fit\\time (s)} & \makecell{Predict\\time (s)} & \makecell{Peak\\memory (MB)} \\
\midrule
\endfirsthead
\toprule
Dataset & Accuracy & \makecell{Fit\\time (s)} & \makecell{Predict\\time (s)} & \makecell{Peak\\memory (MB)} \\
\midrule
\endhead
\midrule
\multicolumn{5}{r}{\footnotesize Continued on next page} \\
\endfoot
\bottomrule
\endlastfoot
ACSF1 & 0.940 & 3.71 & 1.87 & 93.6 \\
Adiac & 0.829 & 1.92 & 0.62 & 204.2 \\
ArrowHead & 0.857 & 0.53 & 0.37 & 21.7 \\
BME & 0.980 & 0.43 & 0.18 & 18.6 \\
Beef & 0.833 & 0.60 & 0.12 & 19.1 \\
BeetleFly & 0.950 & 0.28 & 0.06 & 6.3 \\
BirdChicken & 0.900 & 0.27 & 0.06 & 6.2 \\
CBF & 0.982 & 0.44 & 1.08 & 20.1 \\
Car & 0.917 & 1.03 & 0.32 & 34.1 \\
Chinatown & 0.977 & 0.19 & 0.07 & 2.5 \\
ChlorineConcentration & 0.756 & 2.24 & 7.05 & 290.6 \\
CinCECGTorso & 0.960 & 2.07 & 31.74 & 38.3 \\
Coffee & 1.000 & 0.27 & 0.04 & 7.0 \\
Computers & 0.696 & 5.11 & 3.11 & 265.0 \\
CricketX & 0.803 & 2.64 & 1.14 & 287.2 \\
CricketY & 0.810 & 2.61 & 1.10 & 285.0 \\
CricketZ & 0.805 & 2.61 & 1.16 & 286.7 \\
Crop & 0.750 & 57.16 & 10.03 & 4847.0 \\
DiatomSizeReduction & 0.967 & 0.48 & 0.88 & 6.7 \\
DistalPhalanxOutlineAgeGroup & 0.777 & 1.56 & 0.11 & 201.6 \\
DistalPhalanxOutlineCorrect & 0.793 & 1.10 & 0.11 & 189.9 \\
DistalPhalanxTW & 0.705 & 1.58 & 0.12 & 202.7 \\
ECG200 & 0.890 & 0.34 & 0.05 & 30.6 \\
ECG5000 & 0.946 & 2.08 & 4.76 & 342.8 \\
ECGFiveDays & 0.977 & 0.24 & 0.57 & 6.7 \\
EOGHorizontalSignal & 0.605 & 9.78 & 6.53 & 450.9 \\
EOGVerticalSignal & 0.461 & 9.95 & 7.22 & 398.4 \\
Earthquakes & 0.748 & 1.64 & 0.44 & 137.2 \\
ElectricDevices & 0.766 & 101.64 & 7.70 & 5812.6 \\
EthanolLevel & 0.574 & 17.22 & 12.69 & 699.0 \\
FaceAll & 0.785 & 2.41 & 1.64 & 410.1 \\
FaceFour & 0.943 & 0.59 & 0.31 & 16.3 \\
FacesUCR & 0.947 & 1.22 & 2.42 & 130.8 \\
FiftyWords & 0.837 & 2.66 & 0.98 & 281.2 \\
Fish & 0.989 & 1.78 & 0.74 & 99.8 \\
FordA & 0.959 & 18.98 & 2.80 & 1402.3 \\
FordB & 0.838 & 19.29 & 1.78 & 1420.0 \\
FreezerRegularTrain & 0.992 & 0.65 & 4.03 & 46.0 \\
FreezerSmallTrain & 0.951 & 0.30 & 4.06 & 8.6 \\
GunPoint & 1.000 & 0.31 & 0.13 & 13.3 \\
GunPointAgeSpan & 1.000 & 0.48 & 0.25 & 34.0 \\
GunPointMaleVersusFemale & 0.997 & 0.48 & 0.28 & 33.8 \\
GunPointOldVersusYoung & 0.990 & 0.48 & 0.24 & 34.5 \\
Ham & 0.762 & 0.63 & 0.22 & 32.3 \\
HandOutlines & 0.946 & 60.96 & 15.28 & 1024.0 \\
Haptics & 0.494 & 4.62 & 5.19 & 142.0 \\
Herring & 0.672 & 0.48 & 0.17 & 18.0 \\
HouseTwenty & 0.958 & 2.69 & 3.27 & 38.6 \\
InlineSkate & 0.455 & 5.20 & 14.89 & 100.0 \\
InsectEPGRegularTrain & 1.000 & 1.10 & 1.50 & 41.0 \\
InsectEPGSmallTrain & 0.984 & 0.66 & 1.48 & 11.9 \\
InsectWingbeatSound & 0.637 & 1.75 & 5.29 & 130.1 \\
ItalyPowerDemand & 0.957 & 0.26 & 0.20 & 14.8 \\
LargeKitchenAppliances & 0.792 & 9.74 & 5.21 & 389.3 \\
Lightning2 & 0.754 & 0.52 & 0.20 & 18.9 \\
Lightning7 & 0.781 & 1.28 & 0.35 & 42.1 \\
Mallat & 0.939 & 2.28 & 39.43 & 45.5 \\
Meat & 0.917 & 0.91 & 0.27 & 27.6 \\
MedicalImages & 0.770 & 2.19 & 0.82 & 240.0 \\
MiddlePhalanxOutlineAgeGroup & 0.630 & 1.61 & 0.13 & 181.5 \\
MiddlePhalanxOutlineCorrect & 0.828 & 1.20 & 0.21 & 156.7 \\
MiddlePhalanxTW & 0.578 & 1.88 & 0.15 & 182.6 \\
MixedShapesRegularTrain & 0.973 & 11.79 & 41.68 & 530.6 \\
MixedShapesSmallTrain & 0.962 & 2.92 & 36.88 & 85.3 \\
MoteStrain & 0.958 & 0.21 & 0.55 & 5.6 \\
NonInvasiveFetalECGThorax1 & 0.922 & 30.72 & 21.50 & 1949.3 \\
NonInvasiveFetalECGThorax2 & 0.947 & 29.61 & 22.45 & 1949.9 \\
OSULeaf & 0.967 & 1.89 & 0.93 & 126.6 \\
OliveOil & 0.967 & 0.76 & 0.19 & 13.3 \\
PhalangesOutlinesCorrect & 0.824 & 3.44 & 0.46 & 570.9 \\
Phoneme & 0.353 & 5.25 & 28.32 & 177.5 \\
PigAirwayPressure & 0.909 & 5.10 & 5.86 & 104.9 \\
PigArtPressure & 0.986 & 4.95 & 5.29 & 107.2 \\
PigCVP & 0.952 & 4.51 & 5.53 & 110.3 \\
Plane & 1.000 & 0.80 & 0.16 & 61.0 \\
PowerCons & 0.933 & 0.56 & 0.14 & 54.8 \\
ProximalPhalanxOutlineAgeGroup & 0.859 & 1.60 & 0.19 & 161.6 \\
ProximalPhalanxOutlineCorrect & 0.887 & 1.14 & 0.13 & 139.3 \\
ProximalPhalanxTW & 0.815 & 1.61 & 0.19 & 160.7 \\
RefrigerationDevices & 0.555 & 6.71 & 4.08 & 389.2 \\
Rock & 0.880 & 2.15 & 1.82 & 24.8 \\
ScreenType & 0.469 & 7.04 & 4.45 & 389.1 \\
SemgHandGenderCh2 & 0.930 & 10.86 & 16.07 & 406.5 \\
SemgHandMovementCh2 & 0.629 & 15.85 & 11.93 & 552.2 \\
SemgHandSubjectCh2 & 0.889 & 15.95 & 11.87 & 667.0 \\
ShapeletSim & 1.000 & 0.31 & 0.44 & 6.7 \\
ShapesAll & 0.940 & 5.05 & 2.68 & 422.3 \\
SmallKitchenAppliances & 0.789 & 7.46 & 4.68 & 389.2 \\
SmoothSubspace & 0.973 & 0.71 & 0.07 & 65.3 \\
SonyAIBORobotSurface1 & 0.948 & 0.21 & 0.23 & 5.6 \\
SonyAIBORobotSurface2 & 0.944 & 0.22 & 0.34 & 8.0 \\
StarLightCurves & 0.982 & 21.23 & 147.55 & 1009.7 \\
Strawberry & 0.978 & 1.67 & 0.42 & 197.5 \\
SwedishLeaf & 0.971 & 2.24 & 0.74 & 353.5 \\
Symbols & 0.984 & 0.62 & 3.48 & 14.1 \\
SyntheticControl & 0.993 & 1.30 & 0.25 & 204.1 \\
ToeSegmentation1 & 0.974 & 0.33 & 0.33 & 13.2 \\
ToeSegmentation2 & 0.931 & 0.33 & 0.23 & 11.7 \\
Trace & 1.000 & 1.04 & 0.31 & 63.7 \\
TwoLeadECG & 1.000 & 0.22 & 0.53 & 5.4 \\
TwoPatterns & 0.996 & 4.07 & 4.47 & 736.4 \\
UMD & 0.986 & 0.56 & 0.24 & 23.3 \\
UWaveGestureLibraryAll & 0.959 & 16.87 & 53.94 & 918.2 \\
UWaveGestureLibraryX & 0.836 & 5.29 & 10.38 & 579.0 \\
UWaveGestureLibraryY & 0.769 & 5.37 & 10.63 & 600.1 \\
UWaveGestureLibraryZ & 0.785 & 5.26 & 10.21 & 595.9 \\
Wafer & 0.998 & 2.18 & 4.47 & 359.4 \\
Wine & 0.944 & 0.37 & 0.07 & 8.1 \\
WordSynonyms & 0.726 & 1.82 & 1.33 & 177.1 \\
Worms & 0.753 & 3.95 & 1.00 & 146.4 \\
WormsTwoClass & 0.779 & 3.94 & 1.00 & 146.4 \\
Yoga & 0.921 & 1.20 & 5.76 & 110.8 \\
\end{longtable}

\end{document}